\documentclass[10pt,twocolumn,letterpaper]{article}

\usepackage{iccv}
\usepackage{times}
\usepackage{epsfig}
\usepackage{graphicx}
\usepackage{amsmath}
\usepackage{amssymb}

\usepackage{xcolor}

\usepackage{bbm}

\newcommand{\vect}[1]{\boldsymbol{#1}}
\usepackage[ruled,linesnumbered]{algorithm2e}
\usepackage{multirow}
\usepackage{bbding}
\usepackage{subcaption}

\usepackage{authblk}
\usepackage[breaklinks=true,bookmarks=false]{hyperref}

\iccvfinalcopy 

\def\iccvPaperID{****} 

\ificcvfinal\fi

\begin{document}

\title{Contrastive ACE : Domain Generalization Through Alignment of Causal Mechanisms}


\author[1]{Yunqi Wang}
\author[2]{Furui Liu\thanks{Furui Liu is the corresponding author} }
\author[2]{Zhitang Chen}
\author[3]{Qing Lian}
\author[2]{Shoubo Hu}
\author[2]{Jianye Hao}
\author[1]{Yik-Chung Wu}

\affil[1]{University of Hong Kong}  
\affil[2]{Huawei Noah's Ark Lab}
\affil[3]{Hong Kong University of Science and Technology}



\maketitle
\ificcvfinal\thispagestyle{empty}\fi

\begin{abstract}


Domain generalization aims to learn knowledge invariant across different distributions while semantically meaningful for downstream tasks from multiple source domains, to improve the model's generalization ability on unseen target domains. The fundamental objective is to understand the underlying ''invariance'' behind these observational distributions and such invariance has been shown to have a close connection to causality. While many existing approaches make use of the property that causal features are invariant across domains, we consider the causal invariance of the average causal effect of the features to the labels. This invariance regularizes our training approach in which interventions are performed on features to enforce stability of the causal prediction by the classifier across domains. Our work thus sheds some light on the domain generalization problem by introducing invariance of the mechanisms into the learning process. Experiments on several benchmark datasets demonstrate the performance of the proposed method against SOTAs. 

\end{abstract}

\section{Introduction}
The past decades have witnessed the remarkable success of machine learning, especially deep learning models in solving different problems in various fields. However, the performance guarantee of models is under the assumption that the training and testing data are independent and identically distributed, which can be easily violated in real-world applications since the data-generating processes are usually affected by time, environment, experimental conditions, etc.
As a result, models that work well on training data may perform poorly on new data unseen in the model training stage, and thus restrain their deployment for further applications.
It is of great interest to learn a domain-robust model that can be generalized to the domains beyond source data. 
To this end, researchers proposed the domain generalization problem \cite{blanchard2011generalizing}, which aims at improving the robustness of models on unseen data (i.e., target domain) by learning from several training datasets (i.e., source domains).


The most straightforward domain generalization approach is the leave-one-out strategy, which defines one as the target domain for testing and the rest as source domains for training.
To tackle the challenging problem that no data from target domain is available in training, efforts have been made to extract ''invariance'' from source domains. A natural idea is to frame the network to extract stable features which yield invariant predictions across domains~\cite{li2018domain,sun2016deep,muandet2013domain,vapnik1992principles,xu2020adversarial,arjovsky2019invariant}.
Methods under this branch either enforce consistency of the distributions of latent features across source domains \cite{muandet2013domain,li2018domain}, or minimize the gradients of the classification loss with respect to latent features \cite{arjovsky2019invariant}.

The underlying principle behind these approaches is the postulate that causal features are with certain stability across domains. 
Approaches try to recover these features utilizing the invariance property. However, recent studies show that the causal mechanism, rather than the distribution of features, is stable across domains \cite{rojas2018invariant}. Exploring features by only applying a regularizer to enforce invariance, one may find spurious causal features that are in fact only correlated with the label, leading to instability of the trained models \cite{lu2021nonlinear}. Besides, by incorporating cross-domain mechanism invariance, one is able to recover the causal mechanism as well as causal features, with good interpretability of the contributions of individual features on the task at hand. This also benefits tasks like troubleshooting and identification of important features.

To this end, we tackle the problem of domain generalization from a causal perspective by treating machine learning models as Structural Causal Models (SCM). 
Instead of aligning the distributions of latent features across domains, we propose a novel constraint based on the causal attributions in networks measured by Average Causal Effect (ACE) \cite{montavon2017explaining}. By viewing samples of the same class from different domains as positive pairs and those of different classes as negative pairs, ACE contrastive loss is introduced to regularize the learning procedure and encourage domain-independent attributions of extracted features. The superior experimental results on several benchmark datasets demonstrate the effectiveness of the proposed approach in model generalization compared with several baseline methods and thus show the importance of involving causal attributions in the training of the models.

\section{Related Work}

\subsection{Domain Generalization}

Domain generalization remains a challenging yet important problem that has been investigated by many studies in the literature. The classic way of learning models with good generalization ability is to train feature extractors that can generate invariant representations across different source domains. Various methods have been proposed including naive approaches where a single network is trained by directly aggregating all data from source domains together \cite{li2017deeper}, with a designed structure for more robust performance on data of multi-domain distributions \cite{khosla2012undoing} or modified optimization algorithms which minimize dissimilarity of features between different domains \cite{li2017learning}. Specifically, domain invariant component analysis has been proposed to train models under distribution variations resulted from domain shift \cite{muandet2013domain}. In \cite{ghifary2015domain}, they leverage the maximum mean discrepancy as the measure to guide training on multi-task auto-encoders,  under the principle of aligning source data across domains. 
Some other works \cite{dou2019domain,li2017learning,li2019episodic} have introduced meta-learning with adaptive regularizers to improve generalization ability. By employing Model-Agnostic Meta-Learning or similar strategies in domain generalization, domain-specific gradients have been normalized~\cite{li2017learning,li2019episodic} and models are encouraged to extract features respecting inter-class relationships~\cite{dou2019domain}. Data augmentation, as utilized in various applications, has been also demonstrated to be effective in domain generalization~\cite{nuriel2020permuted,zhang2019unseen,zhou2020learning}. Several attempts have been made to enlarge the support of the distributions in training data such as mixing up or blending data points from different domains~\cite{wang2020heterogeneous,yan2020improve,zhang2019unseen}. Moreover, adversarial data augmentation, as well as several alternatives based on GANs, has also been investigated and shows improvements in addressing domain generalization~\cite{carlucci2019domain,rahman2019multi,volpi2018generalizing}.

To better interpret domain generalization, literature aiming at capturing invariant relations under the structure of causality has emerged. It is argued that causal features with respect to the task, such as shape for classifying objects, are stable and invariant features one wants to learn. However, simply enforcing invariance to train feature extractors without causal considerations, one may obtain only correlated but non-causal features, that are spurious invariant representations for the task. Consider the image classification as an example. A dataset contains a lot of cows on the grass. Feeding them to a model, the grass may also be learned as ``invariant'' representation, but it is not the causal feature for identifying the cow \cite{shen2018causally}. To avoid this, the Invariant Causal Prediction (ICP) is first proposed in~\cite{peters2016causal}. It tries to exploit the invariant property of feature set in causality, in the sense that a structural causal model, as well as the invariant distribution of features, are considered. Several latter studies then made improvements by adding intervention on the target variable and attempt to learn invariant predictors or classifiers ~\cite{magliacane2017domain,subbaswamy2019universal}. By reformulating the optimization problems, Arjovsky et al.~\cite{arjovsky2019invariant} introduce invariant risk minimization to distinguish between spurious correlations and the causal ones, which is then extended to nonlinear settings by~\cite{ahuja2020invariant}. 

\subsection{Causal Neural Network Attribution}

Causal neural network attribution refers to the causal effect of a specific input feature on output prediction in neural networks, which aims to quantify inherent causal influences in machine learning \cite{montavon2017explaining}. Most attribution-based studies \cite{bach2015pixel,montavon2017explaining,smilkov2017smoothgrad} have focused on applying overall functional values to define the contribution of input features, while some other methods leverage the gradients or perturbations with occlusion maps to identify the effect of different features \cite{simonyan2013deep}. However, these types of methods are prone to artifacts, which are unlikely to be measured accurately due to non-identifiability of the errors. Specifically, they can be treated as approaches for estimating individual causal effect, which fail to consider the complicated interactions among neurons and thus result in a biased measure of the importance of the input feature. 

In a recent work \cite{chattopadhyay2019neural}, a new unbiased attribution method, called Average Causal Effect (ACE), has been proposed to calculate causal attribution. This metric is derived based on the first principles of causality \cite{pearl2009causality}. Specifically, structural causal models are leveraged by interpreting the original networks as acyclic graphs where higher layers are generated through a hierarchy of interactions on nodes from lower layers and the operator $do(\dot)$, known as do-calculus, is also used \cite{kocaoglu2017causalgan}. Do-calculus or intervention in causality literature refers to the artificial perturbation on some variables of the system, expecting to measure their causal influences on others. When one applies an intervention to a variable, it is set to the fixed value. Tracking the system under this condition, the distributions of other variables belonging to the system then are called interventional distribution. The causal effect of the intervened variable on others is defined mathematically based on the interventional distributions \cite{pearl2009causality}. In a similar way, ACE is defined as the subtraction between the expectation of the output when a particular input feature is under intervention, and a baseline output when the same feature is uniformly perturbed in a fixed interval of values.


\section{Methods}
A domain is defined as a joint distribution over space $\mathcal{X}\times \mathcal{Y}$, where $\mathcal{X}$ and $\mathcal{Y}$ denote the input and label space, respectively. The training data $\mathbf{D}$ in domain generalization consists of several data sets, each of which contains independent and identically distributed instances sampled from one domain.
A na\"ive way to tackle the distribution shift across domains is to aggregate instances from all domains and conduct model training. Suppose there are $m$ instances in total after combining all source data, written as $\mathbf{D}=\{(\vect{x}_{i}, \vect{y}_i)\}^{m}_{i = 1}$. The corresponding Empirical Risk Minimization (ERM) loss is
\begin{equation}
    \mathcal{L} ( \mathbf{D};\theta,\phi) = \sum^m_{i = 1} \ell \left(g_\phi \left(f_{\theta}(\vect{x}_i)\right), \vect{y}_i\right)
    \label{ori_loss}
\end{equation}
where $\ell$ is an appropriate loss function, $f_{\theta}:\mathcal{X} \to \mathcal{Z}$ denotes an encoding model parameterized by $\theta$ that maps the raw input (image) to a latent feature vector, and $g_{\phi}: \mathcal{Z} \to \mathcal{Y}$ denotes a  model parameterized by $\phi$  that maps the latent feature vector to the output label. $\mathcal{Z}$ is the space of latent features and 
\begin{equation}
    \vect{z}_i =  f_\theta(\vect{x}_i),
\end{equation}
is the encoding of the observation $\vect{x}_i$.

To avoid over-fit to the training domains, several different regularizers have been used in addition to the loss as penalty for reducing domain gaps in the space of latent features ~\cite{muandet2013domain,li2018domain}. Unlike minimizing the cross-domain distance directly in the space of latent features, we provide a novel perspective from causality and impose the invariance on the mechanism for all environments. The basic idea is that the true underlying causal mechanisms that map features to labels are cross-domain invariant. It only depends on class but does not depend on the domain index. For samples in the same class, the true causal mechanism from features to label is similar. However, when the sample is with another class, the mechanism shifts.  We design a quantification of the mechanism and use a contrastive loss to enforce this principle in structure learning. Our framework does not rely on domain labels, similar to ERM that collects samples from multiple domains and aggregates them together as the training data. This is an advantage over most of the domain generalization methods, where the domain indexes are essential for representation learning.  Thus, our method is applicable to improve the model's generalization ability to unseen domains even when all training data is within a single domain.


Some theoretical ties are linking the neural networks and the causal models \cite{chattopadhyay2019neural}. By viewing the domain generalization problem from the causal perspective, one deems all datasets as generated from a typical causal framework, known as Structural Causal Model (SCM). Denote $N(l_1,l_2,\dots,l_T)$ by a network of $l$-layers and $l_t \in L = \{l_1,l_2,\dots,l_T\}$ be the set of neurons in the $t$th layer. For neuron $L$, the set of functions defining causal mechanisms is represented by $\gamma$, and the set of exogenous random variables often considered as unobserved common causes are represented by $U$. The corresponding SCM thus can be expressed as function $f_{\text{SCM}}(L,U,\gamma,P_U)$ with $P_U$ referring to the probability distribution of exogenous random variables in set $U$. By interpreting the network $N(l_1,l_2,\dots,l_T)$ as directed acyclic graphs, SCM constructs a hierarchical model which generates outputs of interactions between nodes from lower layers \cite{chattopadhyay2019neural}.  The flexibility of neural networks also raises confidence in the success of the task of using the neural model to capture the causal mechanism from observational data.  


\begin{figure*}[t]
    \centering
    \includegraphics[width=\textwidth]{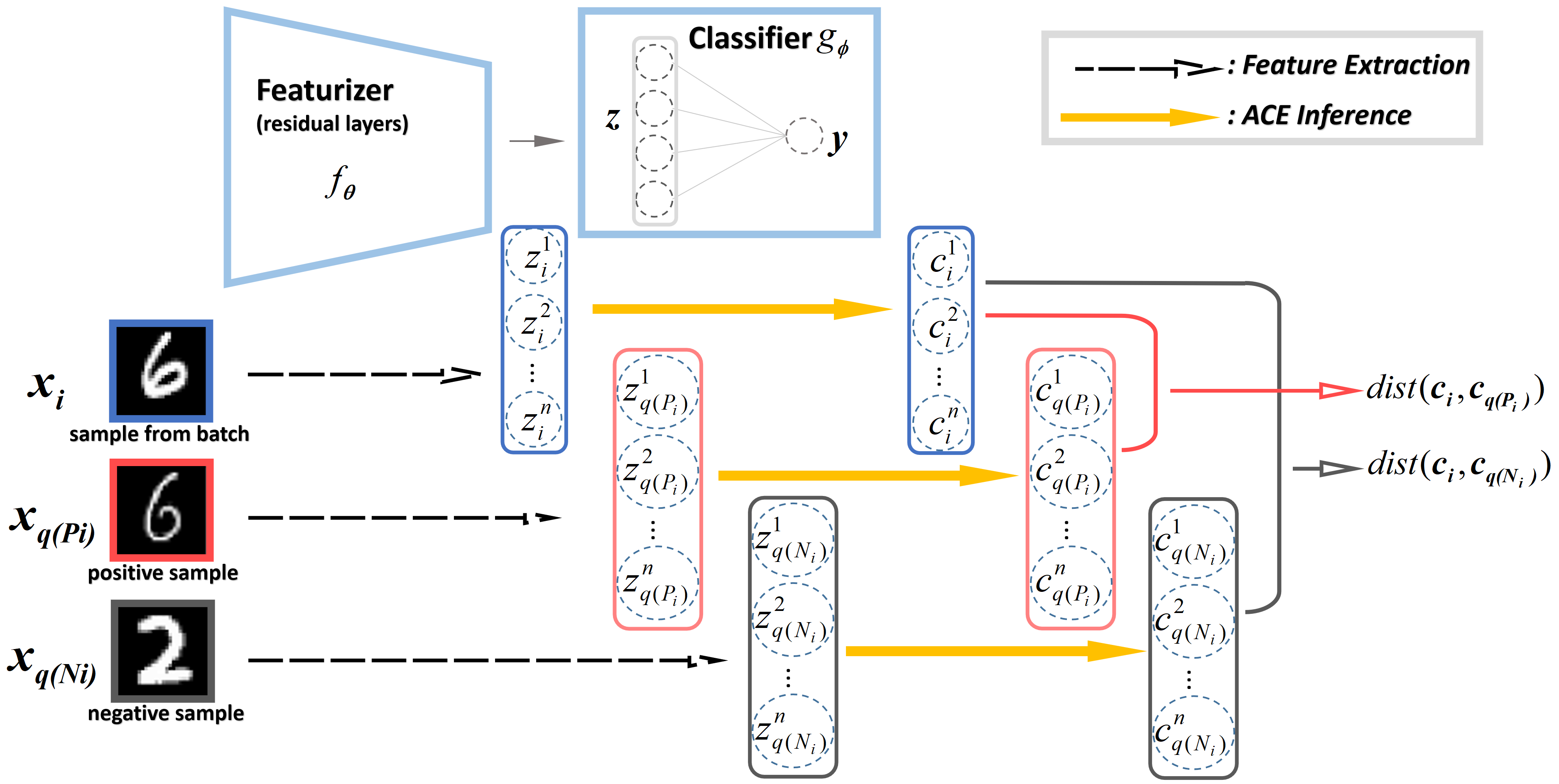}
    \caption{The framework of our method. For each observational image $\vect{x}_i$, it first generates the features by the featurizer (or encoder), implemented by residual neural nets as $\vect{z}_i =  f_\theta(\vect{x}_i)$. $g_\phi$ takes the features to generate the label. The ACE vector $\mathbf{c}_i$ can be computed given the model $g_\phi$,  and the contrastive ACE loss for $\vect{x}_i$ is obtained from the distance between its ACE vector $\mathbf{c}_i$, and the one ($\mathbf{c}_{q(\mathcal{P}_i)}$ or $\mathbf{c}_{q(\mathcal{N}_i)}$) computed using random samples from its positive and negative sets.}
    \label{framework}
\end{figure*}

\subsection{Average Causal Effect}

To identify the causal mechanisms of the task, it is of need to quantify the causal effect of each input feature to the output. Recall that the $f_\theta(\vect{x}_i)$ is an $n$ dimensional vector. Correspondingly, $g_\phi$ is a neural network with $n$ input neurons and an output neuron for each class label. We use $g_\phi$ as the causal quantification model, which  consists of input neurons $\{z^j\}^{n}_{j = 1}$
and output neuron $\vect{y}$. The causal attribution of the neuron $z^j$, corresponding to the $j^{th}$ feature,  on the output $\vect{y}$ is defined as the average causal effect $c_{do(z^j = \alpha)}^{\vect{y}}$ with value $\alpha$ , which can be calculated as the subtraction between interventional expectation of $\vect{y}$ when $z^j = \alpha$ and a baseline of $z^j$ \cite{chattopadhyay2019neural}
\begin{equation}
    c_{do(z^j = \alpha)}^{\vect{y}} = \mathbbm{E}\left[\vect{y}|do(z^j = \alpha)\right] -\mathbbm{E}_{z^j}\left[\mathbbm{E}[\vect{y}|do(z^j = \alpha)]\right].
\end{equation}
The interventional value $\alpha$ can be set to any value in the input domain of $z^j$ as
\begin{equation}
    [\text{low}^{j}, \text{high}^{j}].
\end{equation}
When not intervened, the input neuron $z^j$ is assumed to be uniformly distributed between $\text{low}^{j}$ and $\text{high}^{j}$.
Specifically, the term $\mathbbm{E}[\vect{y}|do(z^j = \alpha)]$, known as the interventional expectation of output neuron $y$ condition on the intervention operation $do(z^j = \alpha)$, is defined as
\begin{equation}
    \mathbbm{E}[\vect{y}|do(z^j = \alpha)] = \int_{\vect{y}} \vect{y} \cdot p(\vect{y}|do(z^j = \alpha)) d\vect{y}.
\end{equation}
The average interventional expectation of $\vect{y}$ with respect to $z^j$, $\mathbbm{E}_{z^j}[\mathbbm{E}[\vect{y}|do(z^j = \alpha)]]$ is used as the baseline value of $z^j$, i.e.
\begin{align}\nonumber
   & \mathbbm{E}_{z^j}[\mathbbm{E}[\vect{y}|do(z^j = \alpha)]]  = \\
    & \int_{\text{low}^j}^{\text{high}^j} p(z^j) \cdot \int_{\vect{y}} \vect{y} \cdot p(\vect{y}|do(z^j = \alpha)) d\vect{y} dz^j,
\end{align}
which has been demonstrated to be unbiased ~\cite{chattopadhyay2019neural}. Hence, the causal attribution of a feature neuron $z^j$ to an output label $\vect{y}$ can be quantified by ACE  $c^{\vect{y}}_{do(z^j = \alpha)}$.

\subsection{Contrastive ACE}
Inspired by the contrastive representation learning \cite{chen2020simple}, we propose a new objective function named contrastive ACE loss, to evaluate the difference between the ACE values of the input feature on output $\vect{y}$ across domains. We first introduce the ACE vector as a quantification of the causal influences of all features on the label of the $i$th sample.
 The feature of the $i$th sample  is a $n$-dimensional vector as
 \begin{equation}
     \vect{z}_i =  f_\theta(\vect{x}_i) = [z_i^1,z_i^2,...,z_i^n].
 \end{equation}
 The ACE of its $j$th feature on the label $\vect{y}$ is defined as
\begin{equation}
    c_i^j = c^{\vect{y}}_{do(z^j = z_i^j)}.
\end{equation}
Going through all dimensions, we get a ACE vector of the $i$th sample 
\begin{equation}
    \vect{c}_i = [c_i^1,c_i^2,...c_i^n].
\end{equation}

\begin{figure}[h]
\includegraphics[width = 8 cm]{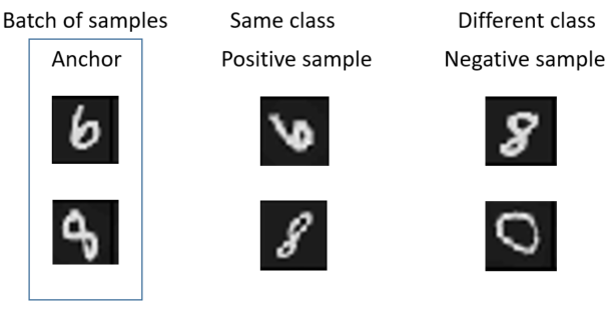}
\caption{Triplet generation: for one observational image $\vect{x}_i$ with label $\vect{y}_i$, its positive set $\mathcal{P}_i$ consists of images that are in the same class as $\vect{y}_i$, and its negative set $\mathcal{N}_i$ consists of images that are with a class different from $\vect{y}_i$.}
\label{p_and_n_sample}
\end{figure}

\begin{figure}[h]
    \centering
    \includegraphics[width=.35\textwidth]{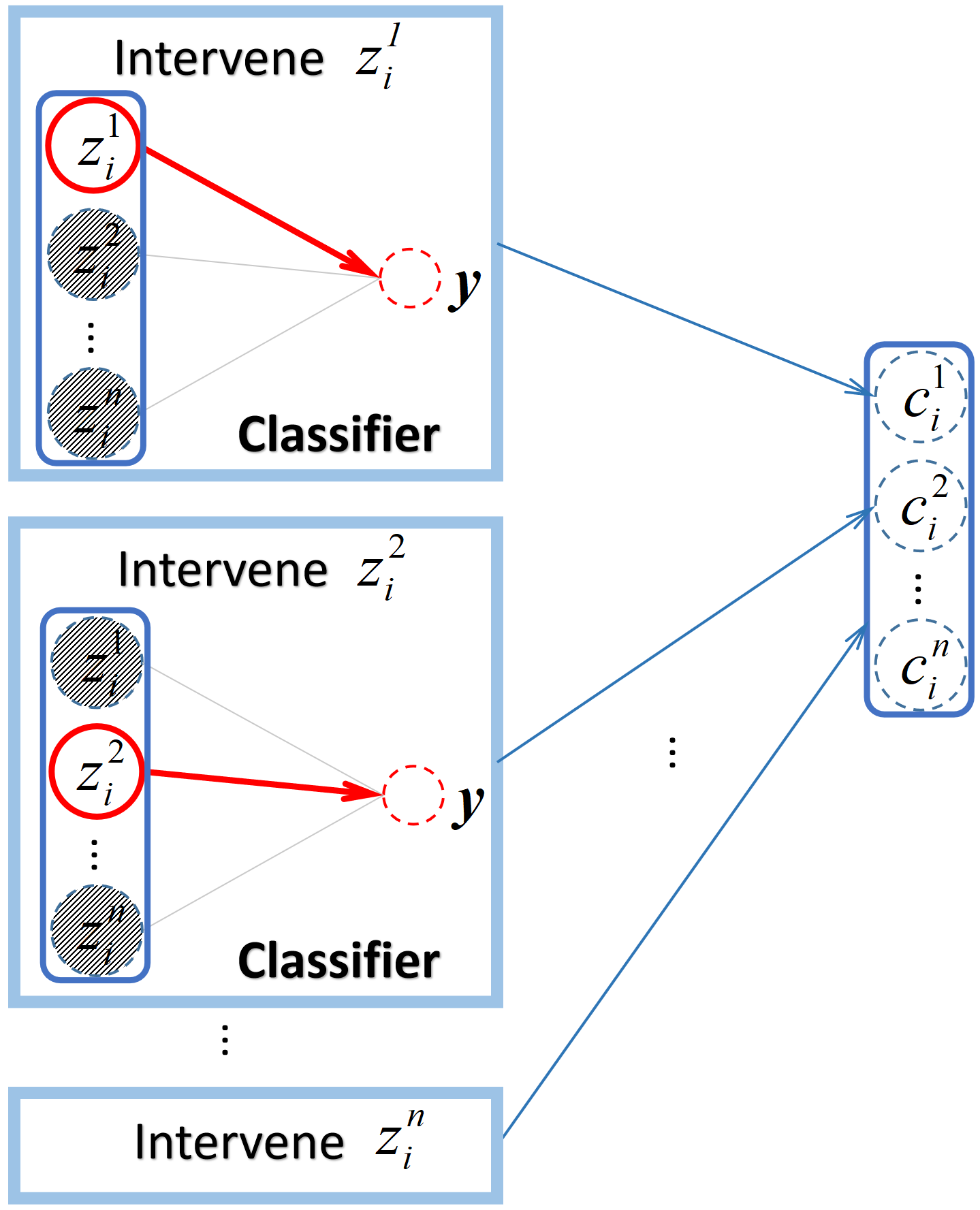}
    \caption{Computation of the ACE vector. Given the features of a sample $\vect{z}_i =  f_\theta(\vect{x}_i)$, the ACE quantification of the classifier $g_\phi$ with $n$ input neurons  $\{z^j\}^{n}_{j = 1}$ is a vector $\vect{c}_i$, which is generated by treating the $j$th neuron intervened as $do(z^j = z_i^j)$. }
    \label{ACE computing}
\end{figure}

An illustration of obtaining the ACE vector is in Fig. \ref{ACE computing}.
To match the ACE of an input to an output for all instances of the same class across domains, the contrastive ACE loss is optimized by minimizing the distance between inputs of the same class and maximizing those from different classes. We treat the inputs of the same class as the positive matches of $i$th sample, and the ones of different classes as negative samples. The positive sets and negative sets for the $i$th sample are
\begin{eqnarray}
\mathcal{P}_i & = & \{k|\vect{y}_i = \vect{y}_k,\forall k\neq i\}, \label{set_p} \\
\mathcal{N}_i & = & \{k|\vect{y}_i \neq \vect{y}_k,\forall k\neq i\} ,
\label{set_n}
\end{eqnarray}
respectively. An intuitive example is also depicted in Fig. \ref{p_and_n_sample}, taking rotated MNIST dataset as an example.

A direct way to measure the overall pairwise distance between ACE vectors is to calculate a distance averaged over all samples in the whole set, which is under extremely heavy computational workloads when the number of samples is large. Thus, we use an efficient random sampling technique.   Denote $q(\mathcal{P}_i)$ an index sampled uniformly at random from the set $\mathcal{P}_i$, and $q(\mathcal{N}_i)$ an index sampled uniformly at random from the set $\mathcal{N}_i$. The contrastive ACE loss $\mathcal{A}_{\theta,\phi}(\vect{x}_i)$ is then defined as
\begin{equation}
\max\left\lbrace dist(\mathbf{c}_i, \mathbf{c}_{q(\mathcal{P}_i)}) - dist(\mathbf{c}_i, \mathbf{c}_{q(\mathcal{N}_i)}) + \delta, 0 \right\rbrace,
\end{equation}
and $\delta > 0$ is a small margin variable \cite{BMVC2016_119}. The loss $\mathcal{A}_{\theta,\phi}(\vect{x}_i)$ becomes large when $dist(\mathbf{c}_i, \mathbf{c}_{q(\mathcal{P}_i)})$ is large, or $dist(\mathbf{c}_i, \mathbf{c}_{q(\mathcal{P}_i)})$ is very small. Thus, it penalizes the inter-class dissimilarity and intra-class similarity. The margin variable here is used to reduce the non-robustness brought by the $\max$ operation, avoiding cases that the $dist(\mathbf{c}_i, \mathbf{c}_{q(\mathcal{P}_i)}) - dist(\mathbf{c}_i, \mathbf{c}_{q(\mathcal{N}_i)})$ is always below 0 and never penalized.  The distance we use is the Manhattan Distance between the pair of vectors as
\begin{equation}
dist(\mathbf{c}_i, \mathbf{c}_j) = |\mathbf{c}_i - \mathbf{c}_j|_M=\sum_{r=1}^n|c_i^r - c_j^r|.
\end{equation}
 Combined with the ERM original loss in Eq. \ref{ori_loss}, our loss with weighting parameter $\rho$ can be written as
 \begin{equation}
         \mathcal{L}_\mathcal{A} ( \mathbf{D};\theta,\phi) = \sum^m_{i = 1} \ell (g_\phi (f_{\theta}(\vect{x}_i)), \vect{y}_i) + \rho \sum^m_{i = 1}  \mathcal{A}_{\theta,\phi}(\vect{x}_i).
         \label{ACE_loss}
 \end{equation}
 
We show the whole training framework in Fig. \ref{framework}, and the pseudo-code of the method in Alg. \ref{pesudo_code}.  Intuitively, our structural loss originates from the principle that the causal mechanism or structural causal model from features to labels is class-dependent, but domain-independent, or cross-domain stable. Given an observation, the ACE vector is a quantification of its features' influence on its label. For samples that are within the same class, we minimize the gap between their quantification vectors; but for samples that are with different classes, a larger gap is preferred. The loss that explicitly addresses the principle is designed in a contrastive way that positive and negative pairs are used.  Incorporating this in the training procedure, we expect our model to recover the true invariant causal structure, which can achieve stable performance in the presence of domain shifts.

\begin{algorithm}[t]
\caption{ACE contrastive learning}
\label{pesudo_code}
\KwIn{ Data $\mathbf{D}=\{(\vect{x}_{i}, \vect{y}_i)\}^{m}_{i = 1}$, parameter $\rho$ and $\delta$. }
\KwOut{Optimal Network. }
Initialize $f_\theta$, $g_\phi$\;
\For{$i=1$ to $m$}{
 Construct $\mathcal{P}_i$ and $\mathcal{N}_i$ as Eq. \ref{set_p} and \ref{set_n} respectively.
}
\While{Not converged}{
Compute the loss as Eq. \ref{ACE_loss}\;
Update $\theta$ and $\phi$ by gradient descent\;
}
\Return $f_\theta$, $g_\phi$;
\end{algorithm}

\section{Experiments}
In this section, we perform experiments to test the performance of our methods on several benchmark datasets, including simulated dataset (Rotated MNIST \cite{ghifary2015domain}) and real-world datasets (PACS \cite{li2017deeper}, VLCS \cite{fang2013unbiased}).  We compare our method with a set of domain generalization approaches. Out of them, ERM is without using the domain indexes, and all other methods take use of the domain indexes. Accuracy is the main metric being compared.



\begin{table*}[t!]
\caption{\upshape Accuracy of Rotated MNIST dataset on target domains from $0^{\circ}$ to $75^{\circ}$. *This result is obtained from normalized MNIST dataset.}
\begin{tabular}{lllllllll}
\hline
\textbf{Method}                 & \textbf{$0^{\circ}$} & \textbf{$15^{\circ}$} & \textbf{$30^{\circ}$} & \textbf{$45^{\circ}$} & \textbf{$60^{\circ}$} & \textbf{$75^{\circ}$} & \textbf{Avg}  & \textbf{Domain Label}         \\ \hline
IRM \cite{arjovsky2019invariant}                             & $96.0 \pm 0.2$ & $98.9 \pm 0.0$  & $99.0 \pm 0.0$  & $98.8 \pm 0.1$  & $98.9 \pm 0.1$  & $95.7 \pm 0.3$  & $97.9$          & \multirow{8}{*}{required}     \\
DRO \cite{sagawa2019distributionally}                             & $96.2 \pm 0.1$ & $98.9 \pm 0.0$  & $99.0 \pm 0.1$  & $98.7 \pm 0.1$  & $99.1 \pm 0.0$  & $96.8 \pm 0.1$  & $98.1$          &                               \\
Mixup \cite{xu2020adversarial}                           & $95.8 \pm 0.3$ & $98.9 \pm 0.1$  & $99.0 \pm 0.1$  & $99.0 \pm 0.1$  & $98.9 \pm 0.1$  & $96.5 \pm 0.1$  & $98.0$          &                               \\
MLDG \cite{wang2020meta}                            & $96.2 \pm 0.1$ & $99.0 \pm 0.0$  & $99.0 \pm 0.1$  & $98.9 \pm 0.1$  & $99.0 \pm 0.1$  & $96.1 \pm 0.2$  & $98.0$          &                               \\
CORAL \cite{sun2016deep}                           & $96.4 \pm 0.1$ & $99.0 \pm 0.0$  & $99.0 \pm 0.1$  & $99.0 \pm 0.0$  & $98.9 \pm 0.1$  & $96.8 \pm 0.2$  & $\textbf{98.2}$ &                               \\
MMD \cite{li2018domain}                             & $95.7 \pm 0.4$ & $98.8 \pm 0.1$  & $98.9 \pm 0.1$  & $98.8 \pm 0.1$  & $99.0 \pm 0.0$  & $96.3 \pm 0.2$  & $97.9$          &                               \\
DANN \cite{ganin2016domain}                            & $96.0 \pm 0.1$ & $98.8 \pm 0.1$  & $98.6 \pm 0.1$  & $98.7 \pm 0.1$  & $98.8 \pm 0.1$  & $96.4 \pm 0.1$  & $97.9$          &                               \\
CDANN \cite{li2018deep}                           & $95.8 \pm 0.2$  & $98.8 \pm 0.0$  & $98.9 \pm 0.0$  & $98.6 \pm 0.1$  & $98.8 \pm 0.1$  & $96.1 \pm 0.2$  & $97.8$          &                               \\ \hline
ERM \cite{vapnik1992principles}                             & $96.0 \pm 0.2$ & $98.8 \pm 0.1$  & $98.8 \pm 0.1$  & $99.0 \pm 0.0$  & $99.0 \pm 0.0$  & $96.8 \pm 0.1$  & $98.1$          & \multirow{3}{*}{not required} \\ \cline{1-8}
\textbf{Contrastive-ACE (ours)} & $96.4 \pm 0.3$       & $98.7 \pm 0.2$        & $99.4 \pm 0.1$        & $99.4 \pm 0.1$        & $99.4 \pm 0.1$        & $97.4 \pm 0.2$        & $\textbf{98.5}$ &                               \\
\textbf{Contrastive-ACE*}         & $97.4 \pm 0.2$       & $99.5 \pm 0.1$        & $99.6 \pm 0.1$        & $99.6 \pm 0.1$        & $99.5 \pm 0.1$        & $98.0 \pm 0.1$        & $\textbf{98.9}$ &                               \\ \hline
\end{tabular}
\label{result_romnist}
\end{table*}

\subsection{Experimental Settings}
We mostly follow the model set up in the paper \cite{gulrajani2020search}.  The models are trained on source domains that are generated from training dataset and evaluated on the target domain which is generated from testing dataset. Source and target domains are generated by the leave-one-out strategy, that one domain is the test and others are as training domains. When image data is the input, the domain generalization models contain the encoder $f_\theta$ and the classifier $g_\phi$. For fair comparisons, all models are with the same  $f_\theta$ and $g_\phi$ as the  DomainBed \cite{gulrajani2020search}.  $f_\theta$ is a 4-layer CNN with residual network and pooling layers (small net for Rotated MNIST and big net for other two datasets), and  $g_\phi$ is a linear FC net.  For model selection, we use the test-domain-validation-set, where a validation set that follows the distribution of the test domain is used to select the best model. The training epoch is fixed to be 100 with batch size $64$.  Adam optimizer is used without weight decay, with a learning rate to be  $0.001$. The hyperparameters of our methods, namely the weight of the ACE regularizer and the margin variable, are $\rho = 1$ and $\delta = 0.05$. The experiments are run 2 times for each dataset, and the average performance, as well as its statistical variation, are reported. 

\begin{figure}[t]
\includegraphics[width=.48\textwidth, height = .2\textwidth]{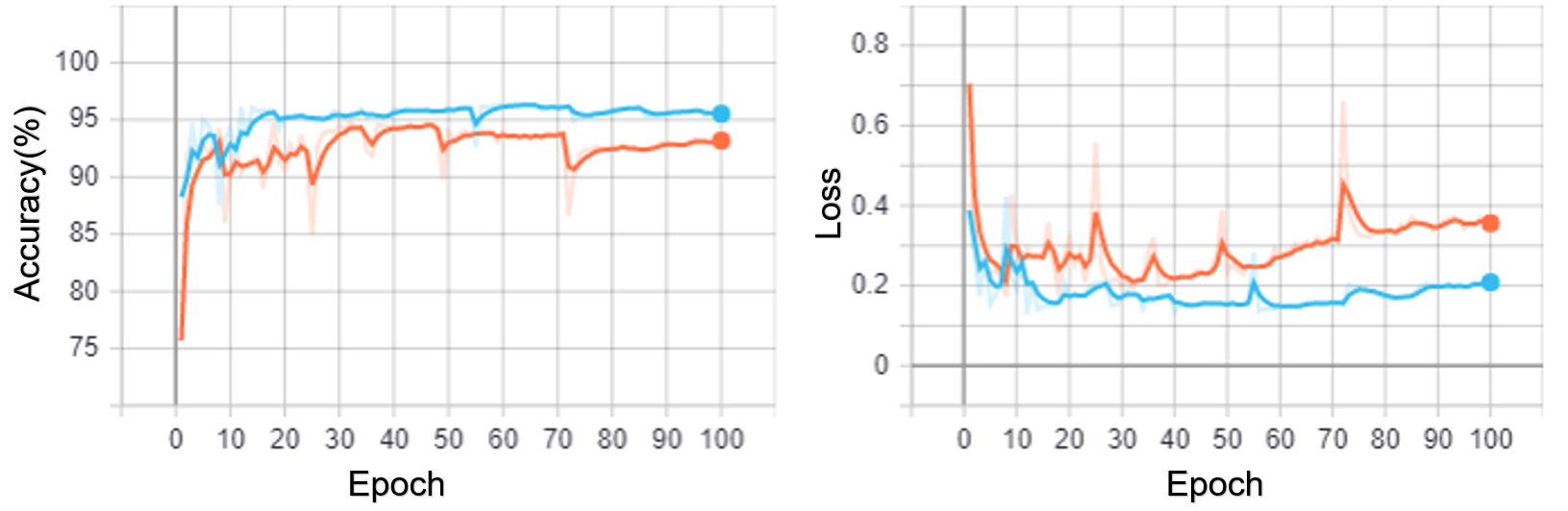}
\caption{Performance on target domain. The orange curve records the historical model accuracy when the training algorithm is ERM, and the blue curve records our results.}
\label{acc_loss_romnist}
\end{figure}

\subsection{Experiments on Rotated MNIST}
This dataset is an artificial dataset constructed from the popular MNIST handwritten digit sets. It contains grayscale MNIST handwritten digits with different rotations, with a degree from $0^{\circ}$ to $75^{\circ}$,  with $15^{\circ}$ as one step interval. The images that are with the same degree of rotation thus naturally form one domain, so that each domain is indexed by the rotation angle.

\begin{figure}[t]
     \centering
     \begin{subfigure}[b]{0.23\textwidth}
         \centering
         \includegraphics[width=\textwidth]{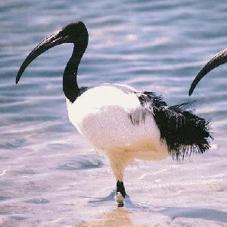}
         \caption{Domain C}
         \label{Cbird}
     \end{subfigure}
     \hfill
     \begin{subfigure}[b]{0.23\textwidth}
         \centering
         \includegraphics[width=\textwidth]{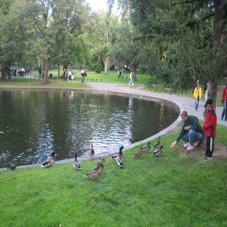}
         \caption{Domain L}
         \label{Lbird}
     \end{subfigure}
     \hfill
     \begin{subfigure}[b]{0.23\textwidth}
         \centering
         \includegraphics[width=\textwidth]{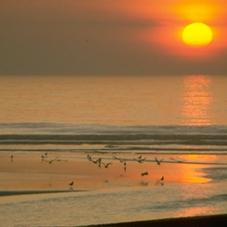}
         \caption{Domain S}
         \label{Sbird}
     \end{subfigure}
     \hfill
     \begin{subfigure}[b]{0.23\textwidth}
         \centering
         \includegraphics[width=\textwidth]{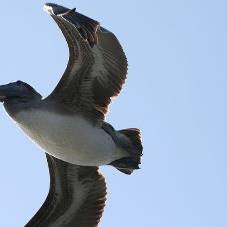}
         \caption{Domain V}
         \label{Vbird}
     \end{subfigure}
     \caption{Four samples of the class "bird" from C, L, S, V four domains. Notice the difference in the proportion of space that the objects occupy in whole the image. }
        \label{vlcs}
\end{figure}

\begin{table*}[t]
\caption{\upshape Model accuracy on VLCS dataset on target domains.}
\centering
\begin{tabular}{lllllll}
\hline
\textbf{Method}     & \textbf{C} & \textbf{L} & \textbf{S} & \textbf{V} & \textbf{Avg}  & \textbf{Domain Label}         \\ \hline
IRM \cite{arjovsky2019invariant}                 & $97.6 \pm 0.5$ & $64.7 \pm 1.1$ & $69.7 \pm 0.5$ & $76.6 \pm 0.7$ & $77.2$          & \multirow{8}{*}{required}                      \\
DRO \cite{sagawa2019distributionally}                 & $97.8 \pm 0.0$ & $66.4 \pm 0.5$ & $68.7 \pm 1.2$ & $76.8 \pm 1.0$ & $77.4$          &                               \\
Mixup \cite{xu2020adversarial}               & $98.3 \pm 0.3$ & $66.7 \pm 0.5$ & $73.3 \pm 1.1$ & $76.3 \pm 0.8$ & $78.7$          &                               \\
MLDG \cite{wang2020meta}                & $98.4 \pm 0.2$ & $65.9 \pm 0.5$ & $70.7 \pm 0.8$ & $76.1 \pm 0.6$ & $77.8$          &                               \\
CORAL\cite{sun2016deep}               & $98.1 \pm 0.1$ & $67.1 \pm 0.8$ & $70.1 \pm 0.6$ & $75.8 \pm 0.5$ & $77.8$          &                               \\
MMD \cite{li2018domain}                 & $98.1 \pm 0.3$ & $66.2 \pm 0.2$ & $70.5 \pm 1.0$ & $77.2 \pm 0.6$ & $78.0$          &                               \\
DANN \cite{ganin2016domain}                & $98.2 \pm 0.3$ & $67.8 \pm 1.1$ & $74.2 \pm 0.7$ & $80.1 \pm 0.6$ & $80.1$          &                               \\
CDANN \cite{li2018deep}               & $98.9 \pm 0.3$ & $68.8 \pm 0.6$ & $73.7 \pm 0.6$ & $79.3 \pm 0.6$ & $\textbf{80.2}$ &                               \\ \hline
ERM \cite{vapnik1992principles}                 & $97.7 \pm 0.3$ & $65.2 \pm 0.4$ & $73.2 \pm 0.7$ & $75.2 \pm 0.4$ & $77.8$          & \multirow{2}{*}{not required} \\ \cline{1-6}
\textbf{Contrastive-ACE (ours)} & $98.5 \pm 0.5$ & $65.1 \pm 0.3$ & $70.9 \pm 1.2$ & $77.7 \pm 0.6$ & $\textbf{78.1}$ &                               \\ \hline
\end{tabular}
\label{result_vlcs}
\end{table*}

As reported in Table \ref{result_romnist}, we obtain an average accuracy of $98.5\%$, which is the best among all other approaches. An interesting observation is that the approaches without the need for domain labels are, in general, with better performance compared to the ones that need domain labels. When the domain of 0 rotation is used as the testing domains, methods, in general, perform slightly worse than other settings. Recently, debating about the role of normalization emerges \cite{li2020feature} and we also explore its effect on the performance of our approaches. With a simple mean-std normalization of the data, we find that our ACE-based approach achieves a higher accuracy of $98.9\%$. This is possibly because that the normalized data is with a more stable range, which is less sensitive to additive noises and thus with a ground for making better ACE estimation and recovery of the causal mechanism.


To make an in-depth analysis of the training procedure, we plot the accuracy on testing domains of the models trained by ERM and ACE in Fig. \ref{acc_loss_romnist}. We observe that in the first 10 training epochs,  ERM performs much similar to ours, with no obvious difference in-between. Our model clearly outperforms ERM as the training proceeds. This is because that in the initial exploration stage, the structure of neural networks are unstable, and the causal mechanism and features are not recovered to a satisfactory degree. However, when it approaches a relatively mature stage, ACE takes its effect to help guide the model to discover causal features so that stable performance is achieved.

\subsection{Experiments on VLCS}
As one of the classic benchmark datasets for domain generalization, VLCS collects natural images from four datasets, i.e. PASCAL VOC2007 (V), LabelMe (L), Caltech (C), and SUN09 (S), and contains a total of five classes for recognition task (bird, car, chair, dog and person). The images in VLCS are all collected from the real world, which have larger intra-class variance and significantly higher domain shift compared to the simulated datset such as Rotated MNIST. The task of domain generalization thus becomes much more challenging.


As reported in Table \ref{result_vlcs}, we achieve an average classification accuracy of $78.1\%$, which is the best among the approaches that do not require domain labels. We observe a huge difference in the performance across different domains (from $98.5\%$ to $65.1\%$), which indicates the large distribution shift across domains. As opposed to Rotated MNIST, the approaches that require domain labels perform generally better than those do not in VLCS dataset. It is in accordance with natural intuition that domain information brought by labels makes much more critical impact when handling datasets of larger distribution shifts. Compared with the increase in accuracy of ERM ($+0.4\%$) in Rotated MNIST, the improvement of aligning causal mechanism in VLCS is diminished. As complex real-world images contain more complicated and diverse features than simulated images, it is more difficult to infer the causal relationship between features and predictions, despite the presence of the contrastive ACE penalty.


\begin{figure}
     \centering
     \begin{subfigure}[b]{0.23\textwidth}
         \centering
         \includegraphics[width=\textwidth]{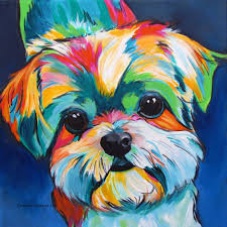}
         \caption{Domain A}
         \label{Adog}
     \end{subfigure}
     \hfill
     \begin{subfigure}[b]{0.23\textwidth}
         \centering
         \includegraphics[width=\textwidth]{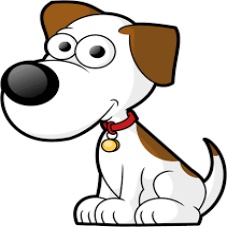}
         \caption{Domain C}
         \label{Cdog}
     \end{subfigure}
     \hfill
     \begin{subfigure}[b]{0.23\textwidth}
         \centering
         \includegraphics[width=\textwidth]{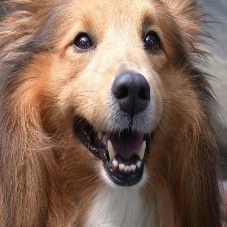}
         \caption{Domain P}
         \label{Pdog}
     \end{subfigure}
     \hfill
     \begin{subfigure}[b]{0.23\textwidth}
         \centering
         \includegraphics[width=\textwidth]{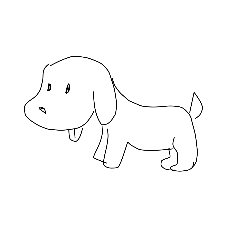}
         \caption{Domain S}
         \label{Sdog}
     \end{subfigure}
     \caption{Four samples of the class "dog" from A, C, P, S four domains. The objects in four domains
     all occupy a large proportion in the image and are well centralized. The distribution shift across domains attributes to the completely style shifts. }
        \label{pacs}
\end{figure}


\begin{table*}[t]
\centering
\caption{\upshape Model accuracy  on PACS dataset on target domains.}
\begin{tabular}{lllllll}
\hline
\textbf{Method}     & \textbf{A} & \textbf{C} & \textbf{P} & \textbf{S} & \textbf{Avg}  & \textbf{Domain Label}         \\ \hline
IRM\cite{arjovsky2019invariant}                 & $85.7 \pm 1.0$ & $79.3 \pm 1.1$ & $97.6 \pm 0.4$ & $75.9 \pm 1.0$ & $84.6$          & \multirow{8}{*}{required}     \\
DRO \cite{sagawa2019distributionally}                 & $88.2 \pm 0.7$ & $82.4 \pm 0.8$ & $97.7 \pm 0.2$ & $80.6 \pm 0.9$ & $\textbf{87.2}$          &                               \\
Mixup \cite{xu2020adversarial}               & $87.4 \pm 1.0$ & $80.7 \pm 1.0$ & $97.9 \pm 0.2$ & $79.7 \pm 1.0$ & $86.4$          &                               \\
MLDG \cite{wang2020meta}                & $87.1 \pm 0.9$ & $81.3 \pm 1.5$ & $97.6 \pm 0.4$ & $81.2 \pm 1.0$ & $86.8$          &                               \\
CORAL \cite{sun2016deep}              & $87.4 \pm 0.6$ & $82.2 \pm 0.3$ & $97.6 \pm 0.1$ & $80.2 \pm 0.4$ & $86.9$          &                               \\
MMD  \cite{li2018domain}               & $87.6 \pm 1.2$ & $83.0 \pm 0.4$ & $97.8 \pm 0.1$ & $80.1 \pm 1.0$ & $87.1$          &                               \\
DANN  \cite{ganin2016domain}               & $86.4 \pm 1.4$ & $80.6 \pm 1.0$ & $97.7 \pm 0.2$ & $77.1 \pm 1.3$ & $85.5$          &                               \\
CDANN \cite{li2018deep}              & $87.0 \pm 1.2$ & $80.8 \pm 0.9$ & $97.4 \pm 0.5$ & $77.6 \pm 0.1$ & $85.7$ &                               \\ \hline
ERM \cite{vapnik1992principles}                 & $87.8 \pm 0.4$ & $82.8 \pm 0.5$ & $97.6 \pm 0.4$ & $80.4 \pm 0.6$ & $87.2$          & \multirow{2}{*}{not required} \\ \cline{1-6}
\textbf{Contrastive-ACE (ours)}  & $88.8 \pm 1.3$ & $81.9 \pm 1.2$ & $97.7 \pm 0.2$ & $80.6 \pm 0.3$ & $\textbf{87.3}$ &                               \\ \hline
\end{tabular}
\label{result_pacs}
\end{table*}

\subsection{Experiments  on PACS}

PACS dataset has recently been widely adopted as a benchmark dataset for domain generalization, which is even more challenging than VLCS. A total $7$ classes of images (dog, elephant, giraffe, guitar, house, horse, and person) from $4$ different domains (art painting, cartoon, photo, and sketch) are included. PACS is considered to have a significantly higher domain shift than VLCS, which attributes to the large difference in style. As shown in Figure \ref{pacs}, the objects in PACS dataset are better positioned by taking a large portion of the image and well centralized compared to those in VLCS dataset. 

As reported in Table \ref{result_pacs}, we achieve an average classification accuracy of $87.3\%$, which is the best among all the methods, including the ones that use domain labels. It is interesting to observe that, although the inter-source domain divergence in PACS dataset is considered to be larger than that in VLCS dataset \cite{wang2020learning} \cite{pandey2020discrepancy}, the difference in the model's generalization performance is relatively small compared to VLCS dataset. This is because the representations extracted by the featurizer contain more information related to the object rather than from the background. For example, a large proportion of features from Figure \ref{Sbird} might contain patterns depicting the sun and the sea while neglecting the bird object. On the contrary, the featurizer may be easier to extract causal features from the dog in Figure \ref{Cdog}. 

Model's performance when domain P is used for testing and others for training are superior compared to other settings, owing to the featurizer realized by the backbone of ResNet-50 pretrained on ImageNet, which is comprised of real-world photos from domain P. The bias brought by featurizer (the pretrained backbone), still influences the generalization capability of the domain models. Although reduced, this bias cannot be completely removed even we use ACE contrastive learning.  It remains a challenging problem that hasn't been well addressed in  existing literatures.

Methods that impose cross-domain invariant representations aim at extracting features that contain domain-independent information while eliminating domain-specific information like styles. However, simply enforcing invariance, one may obtain ''over-fixed'' patterns without cross-domain flexibility that is beneficial for classification.  This leads to inferior classification accuracies of their methods. Instead of enforcing learning domain-invariant features, the proposed contrastive ACE aligns the causal mechanism quantified by the ACE of latent representations to predictions, with room for reasonable variations among patterns. This is possibly the reason that we get better features for the task at hand.

\section{Conclusion}
In this paper, we provide a novel perspective on domain generalization by making use of the causal invariance between the average causal effect of the latent representations to the labels. By assuming the mechanism to be label-dependent but domain-independent, we align causal quantification vectors of samples. A novel contrastive ACE loss is introduced into the training to enforce cross-domain stability in predictions. Without using domain labels, our method still achieves good performance on benchmark datasets compared to SOTAs. The feasibility and effectiveness are demonstrated by extensive experiments. To the best of our knowledge, this work presents the first investigation on aligning causal mechanisms across domains in the learning process to address domain generalization.  We expect that it can motivate researchers to explore along this research line.

{\small
\bibliographystyle{ieee_fullname}
\bibliography{egbib}
}

\end{document}